\documentclass[a4paper,fleqn]{cas-dc}

\usepackage[authoryear,longnamesfirst]{natbib}
\usepackage[utf8]{inputenc}
\usepackage{CJKutf8}
\usepackage{amsmath}
\newcommand{\ourmethod}{EXAM}
\newcommand{\oureval}{SEE}

\newcommand{\zh}[1]{\protect\begin{CJK*}{UTF8}{gbsn}#1\protect\end{CJK*}}

\def\tsc#1{\csdef{#1}{\textsc{\lowercase{#1}}\xspace}}
\tsc{WGM}
\tsc{QE}
\tsc{EP}
\tsc{PMS}
\tsc{BEC}
\tsc{DE}

\begin{document}
\let\WriteBookmarks\relax
\def\floatpagepagefraction{1}
\def\textpagefraction{.001}

\shorttitle{Rethinking the Roles of Large Language Models in Chinese Grammatical Error Correction}

\shortauthors{Yinghui Li et~al.}

\title [mode = title]{Rethinking the Roles of Large Language Models in Chinese Grammatical Error Correction}                      



%

\author[1]{Yinghui Li}[orcid=0000-0001-7571-6722]
\cormark[1]
\ead{liyinghu20@mails.tsinghua.edu.cn}
\credit{ Conceptualization, Formal analysis, Investigation, Methodology, Writing - original draft}

\author[1]{Shang Qin}[orcid=0009-0004-8612-4078]
\cormark[1]
\ead{qin-s23@mails.tsinghua.edu.cn}
\credit{Conceptualization, Formal analysis, Software, Methodology}

\author[1]{Haojing Huang}[orcid=0000-0002-6299-5467]
\cormark[1]
\ead{hhj23@mails.tsinghua.edu.cn}
\credit{Resources, Formal analysis, Validation, Visualization, Writing - original draft}

\author[1]{Yangning Li}[orcid=0000-0002-1991-6698]
\ead{yn-li23@mails.tsinghua.edu.cn}
\credit{Investigation, Data curation}

\author[2]{Libo Qin}[orcid=0000-0002-3619-675X]
\ead{lbqin@csu.edu.cn}
\credit{Writing - review \& editing}

\author[3]{Xuming Hu}[orcid=0000-0001-6075-4224]
\ead{xuminghu@hkust-gz.edu.cn}
\credit{Writing - review \& editing}

\author[4]{Wenhao Jiang}[orcid=0000-0002-0795-366X]
\cormark[2]
\ead{jiangwenhao@gml.ac.cn}
\credit{Project administration, Writing - review \& editing}

\author[1]{Hai-Tao Zheng}[orcid=0000-0001-5128-5649]
\cormark[2]
\ead{zheng.haitao@sz.tsinghua.edu.cn}
\credit{Funding acquisition, Writing - review \& editing}

\author[5]{Philip S. Yu}[orcid=0000-0002-3491-5968]
\ead{psyu@uic.edu}
\credit{Writing - review \& editing}





\affiliation[1]{organization={Tsinghua Shenzhen International Graduate School, Tsinghua
University},
    city={Shenzhen},
    postcode={518055}, 
    state={Guangdong},
    country={China}}
\affiliation[2]{organization={School of Computer Science and Engineering, Central South University},
    city={Changsha},
    postcode={410083}, 
    state={Hunan},
    country={China}}
\affiliation[3]{organization={The Hong Kong University of Science and Technology (Guangzhou)},
    city={Guangzhou},
    postcode={511453}, 
    state={Guangdong},
    country={China}}
\affiliation[4]{organization={Guangdong Laboratory of Artificial Intelligence and Digital Economy (SZ)},
    city={Shenzhen},
    postcode={518132}, 
    state={Guangdong},
    country={China}}
\affiliation[5]{organization={University of Illinois Chicago},
    city={Chicago},
    postcode={60607}, 
    state={Illinois},
    country={USA}}






\cortext[cor1]{indicates equal contribution.}
\cortext[cor2]{Corresponding authors.}



\begin{abstract}
Recently, Large Language Models (LLMs) have been widely studied by researchers for their roles in various downstream NLP tasks. 
As a fundamental task in the NLP field, Chinese Grammatical Error Correction (CGEC) aims to correct all potential grammatical errors in the input sentences. 
Previous studies have shown that LLMs' performance as correctors on CGEC remains unsatisfactory due to the challenging nature of the task.
To promote the CGEC field to better adapt to the era of LLMs, we rethink the roles of LLMs in the CGEC task so that they can be better utilized and explored in CGEC. 
Considering the rich grammatical knowledge stored in LLMs and their powerful semantic understanding capabilities, we utilize LLMs as explainers to provide explanation information to the CGEC small models during error correction, aiming to enhance performance.
We also use LLMs as evaluators to bring more reasonable CGEC evaluations, thus alleviating the troubles caused by the subjectivity of the CGEC task.
In particular, our work is also an active exploration of how LLMs and small models better collaborate in downstream tasks.
Extensive experiments~\footnote{Our code will be made public after peer review.} and detailed analyses on widely used datasets verify the effectiveness of our intuition and the proposed methods.
\end{abstract}


\begin{highlights}
\item LLMs are utilized as explainers to provide auxiliary information during CGEC.
\item We employ LLMs as evaluators to achieve more objective CGEC assessments.
\item Our work actively explores how LLMs and small models collaborate in downstream tasks.
\item Extensive experiments demonstrate the effectiveness of our methods on CGEC datasets.
\end{highlights}


\maketitle

\section{Introduction}
Large Language Models (LLMs) are undoubtedly the hottest topic in the AI and NLP community. Due to the unified paradigm for various tasks and amazing emergent ability, an increasing number of researchers and works have begun to focus on how to better apply LLMs to downstream task scenarios, such as sequence understanding~\cite{yu2023seqgpt}, financial analysis~\cite{wu2023bloomberggpt}, and medical healthcare~\cite{wang2023huatuo}.  
In the vast field of Chinese NLP research, Chinese Grammatical Error Correction (CGEC) has long been regarded as a fundamental task~\cite{DBLP:conf/emnlp/MaLSZHZLLLCZS22}. The CGEC task aims to correct all possible grammatical errors in the input sentence, which is challenging because it requires the models to have a comprehensive understanding ability for the complex semantics of the text. In the era of LLMs, some works have explored the possibility of LLMs for CGEC~\cite{DBLP:journals/corr/abs-2304-01746,li2023effectiveness}. Their consensus is that even with supervised fine-tuning on CGEC data, the performance of LLMs on the CGEC task is still unsatisfactory.
The main reason is that the relatively free generation paradigm makes the sentences generated by LLMs often unable to meet the minimum change principle pursued by CGEC. 
Therefore, adapting and applying LLMs in the CGEC field have encountered a stagnant dilemma.

\begin{figure}[]
\centering
\includegraphics[height=0.33\textwidth]{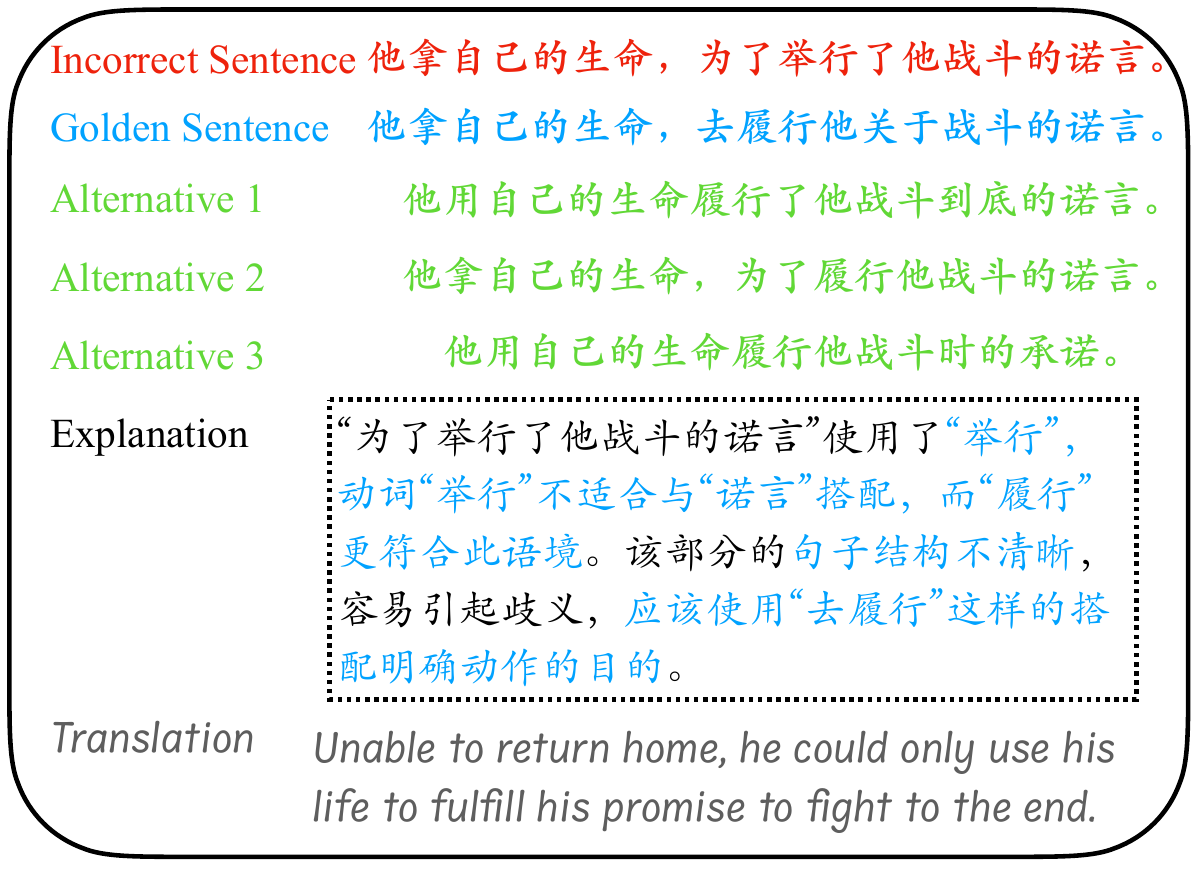}
\caption{The example of subjectivity and explainability of CGEC. The explanation is produced by ChatGPT.}
\label{fig:intro}
\end{figure}

To address this dilemma, our work rethinks the proper utilization of LLMs to promote the development of the CGEC field.
Overviewing recent GEC research trends, the subjectivity and explainability of GEC have received great attention~\cite{ye-etal-2023-cleme, DBLP:journals/corr/abs-2311-09517, DBLP:journals/corr/abs-2309-11439}.
As illustrated in Figure~\ref{fig:intro}, a grammatically incorrect sentence often has different correction methods to keep its meaning unchanged and its grammar correct. Therefore, enabling evaluators to perform comprehensively and flexibly has always been an unsolved challenge.
In addition, we also see from Figure~\ref{fig:intro} that the explanation of the incorrect sentence contains instructive information and knowledge for error correction. If we can obtain high-quality explanations of incorrect sentences, it will undoubtedly improve the CGEC performance.
The basis for high-quality explanations of ungrammatical sentences is rich grammatical knowledge, while flexible CGEC evaluation requires the evaluator to have comprehensive semantic understanding capabilities. 
Intuitively, for LLMs, the massive training corpus gives them \textbf{sufficient grammatical knowledge}, and the emergence phenomenon gives them \textbf{excellent semantic understanding capabilities.} More importantly, the two processes of explanation and evaluation are not restricted by the minimum change principle, and they can give enough free space to the generation paradigm of LLMs.

Motivated by the above intuitions, we believe that LLMs can be leveraged to provide high-quality explanations and accurate evaluations for small CGEC models. Therefore, we propose an \textbf{EX}planation-\textbf{A}ug\textbf{M}ented training framework (\textbf{\ourmethod{}}) and a \textbf{SE}mantic-incorporated \textbf{E}valuation framework (\textbf{\oureval{}}) for CGEC based on LLMs. Specifically, (1) \ourmethod{} mines broad explanation information (including error types, reference corrections, and error explanations) related to grammatically incorrect sentences from LLMs, and then utilizes mined information to enhance the training of small models, thereby ultimately improving the CGEC performance of small models. 
(2) \oureval{} requires LLMs to balance the edits annotated in the golden data with the evaluated model's edits, ensuring they do not alter the original semantics of the input sentence. This ensures more accurate and comprehensive evaluation results that consider both grammar and semantics.
Extensive experiments and detailed analyses demonstrate the effectiveness and competitiveness of our proposed methods. In summary, our technical contributions and impacts are in four folds:
\begin{itemize}
    \item We propose \oureval{}, which aims to empower the evaluation of more subjective CGEC tasks through the intervention of LLMs.
    \item We propose \ourmethod{}, which utilizes LLMs as explainers to enhance the training of small models. This approach enables small models not only to surpass LLMs on traditional metrics but also to demonstrate competitive performance under our proposed \oureval{}.
    \item \textbf{For CGEC field}, we reposition the roles of LLMs to give full play to the strengths of LLMs and promote the adaptation of LLMs to the CGEC task.
    \item \textbf{For LLMs community}, our work explores collaborative cooperation between LLMs and small models on downstream tasks and, to a certain extent, reveals how LLMs and small models can coexist and thrive in the future.
\end{itemize}

\section{Related Work}
In the era of LLMs, considering the superior performance of LLMs~\cite{DBLP:journals/patterns/LiuLTLZ22, DBLP:journals/csur/DongLGCLSY23, DBLP:journals/corr/abs-2303-13547, DBLP:journals/corr/abs-2308-06966,  DBLP:journals/corr/abs-2308-10855, DBLP:journals/corr/abs-2311-11268}, researchers have invested lots of energy in studying LLMs for GEC tasks~\cite{DBLP:conf/acl/LiZLLLSWLCZ22, DBLP:conf/emnlp/LiMZLLHLLC022, DBLP:conf/icassp/ZhangLZMLCZ23, DBLP:journals/corr/abs-2210-12692, DBLP:journals/corr/abs-2306-17447}.

First, some works evaluate LLMs on GEC~\cite{DBLP:journals/corr/abs-2304-01746, DBLP:journals/corr/abs-2306-15788, DBLP:journals/corr/abs-2307-03972, li2023effectiveness, DBLP:conf/wanlp/KwonBNA23, DBLP:conf/emnlp/YeLL023, DBLP:conf/emnlp/HuangYZLLZZ23, DBLP:journals/corr/abs-2401-07702}. In general, GEC-related tasks are challenging for LLMs. There are many reasons for this challenge, such as the inconvenience caused to LLMs by the minimum change principle. 
To address the challenges, some researchers also focus on training LLMs on GEC data~\cite{DBLP:conf/nlpcc/FanJLL23, DBLP:journals/corr/abs-2305-13225, DBLP:conf/nlpcc/SuZQZYZZM23}. Still unsatisfactory, even after supervised fine-tuning, the performance of LLMs still cannot prove that LLMs have fully adapted to the GEC field. For example, the $\text{F}_{0.5}$ scores reported by GrammarGPT~\cite{DBLP:conf/nlpcc/FanJLL23} still do not exceed 40.0.
As a result, researchers begin to pay attention to whether LLMs can have other roles in the GEC field, instead of directly acting as the corrector. 
~\citet{DBLP:conf/emnlp/KanekoO23} propose to improve the GEC performance by letting LLMs predict edit spans.
~\citet{DBLP:journals/corr/abs-2308-08982} and~\citet{DBLP:conf/emnlp/SottanaLZY23} explore the potential of using LLMs as evaluators for English and Swedish GEC tasks.
~\citet{DBLP:journals/corr/abs-2311-09517} and~\citet{DBLP:journals/corr/abs-2309-11439} propose the new task of grammar error explanation and have proved the ability of LLMs to explain grammatical error. 
However, they do not go further to utilize the explanation information in training GEC models.
\textit{To the best of our knowledge, our work is the first to comprehensively think about and design how to make full use of LLMs in the training and evaluation process of GEC small models.
More importantly, our work rethinks how LLMs and small models should coexist and progress together in the era of LLMs, contributing their respective strengths to the advancement of downstream tasks.}

\section{Motivation and Methodology}

\subsection{Motivation}
\paragraph{Minimum Change Principle} 
In the long-term GEC or CGEC research, the setting followed by researchers is the ``minimum change principle'', that is, an ideal model should be able to convert grammatically incorrect sentences into correct sentences with minimal changes or editing costs. 
However, with the development of deep learning and Pre-trained Language Models, the enhancement of model capabilities has conflicted with this principle because it limits the model's space for self-development to a certain extent. Especially with the emergence of LLMs, the performance obtained by directly using LLMs to complete the GEC task is not satisfactory. Many observations and empirical results indicate that the key reason for the unsatisfactory performance of LLMs on CGEC is that the relatively freer text generation mode of LLMs is unsuitable for the GEC task. For example, LLMs often produce sentences that are grammatically correct and semantically consistent with the erroneous input sentence, but the literal text differs significantly from the input sentence. This situation often fails in traditional evaluation metrics, resulting in the low performance of LLMs.

\paragraph{LLMs as Explainer} Given the limitations of directly employing LLMs as correctors due to the minimum change principle, can we adopt an alternative approach to leverage LLMs more effectively for CGEC and circumvent the constraints imposed by this principle? 
First, let's consider what humans do when they encounter grammatical errors, particularly when they are unsure how to correct them.
The most direct and effective solution is to turn to a teacher or grammar reference book. Then, the teacher or reference book would give specific explanations or reasons for grammatical errors to help humans make corrections successfully.
\textbf{Drawing inspiration from human actions, why can't we consider LLMs as explainers similar to teachers or reference books?}
As mentioned in the previous paragraph, the fact that LLMs can generate grammatically correct sentences means that LLMs store rich grammatical knowledge. 
Therefore, we believe that if explanations related to error sentences can be obtained from LLMs and utilized in the training of small models, then these explanations embodying grammatical knowledge from LLMs can definitely enhance the performance of small models. In particular, the role of LLMs as explainers does not need to be limited by the minimum change principle, and it is a simple yet effective process for LLMs to use their own grammatical knowledge to explain incorrect sentences.

\paragraph{LLMs as Evaluator} Considering the subjective nature of the CGEC task, a sentence with grammatical errors often has different correction methods. We argue that the ideal evaluation that can truly reflect the CGEC performance should consider the correction results given by the model as comprehensively as possible. As long as the model provides a sentence that is consistent with the original meaning of the incorrect sentence and has no grammatical errors, its correction should be considered successful. Suppose we want to achieve this ideal evaluation from the perspective of dataset construction. In that case, we need to manually annotate the dataset with as many correct reference sentences corresponding to the incorrect sentences as possible. However, such an annotation process is expensive and time-consuming. Even though there are already multi-reference datasets such as MuCGEC~\cite{DBLP:conf/naacl/0004LBLZLHZ22}, we still believe that automatic evaluation based on such datasets is not flexible enough because the fixed reference correct sentences of the dataset are still limited after all. \textbf{Motivated by the process of teachers correcting students' sentences with grammatical errors, why can't we utilize LLMs as evaluators to play the role of a teacher reviewing grammatical errors?} Intuitively, LLMs not only store rich grammatical knowledge but also have an excellent ability to perceive text semantics. Therefore, we believe that they are fully qualified to be flexible and excellent teachers (i.e., evaluators) who review the answers of models in the GEC task.

\begin{figure*}[]
\centering
\includegraphics[width=0.80\textwidth]{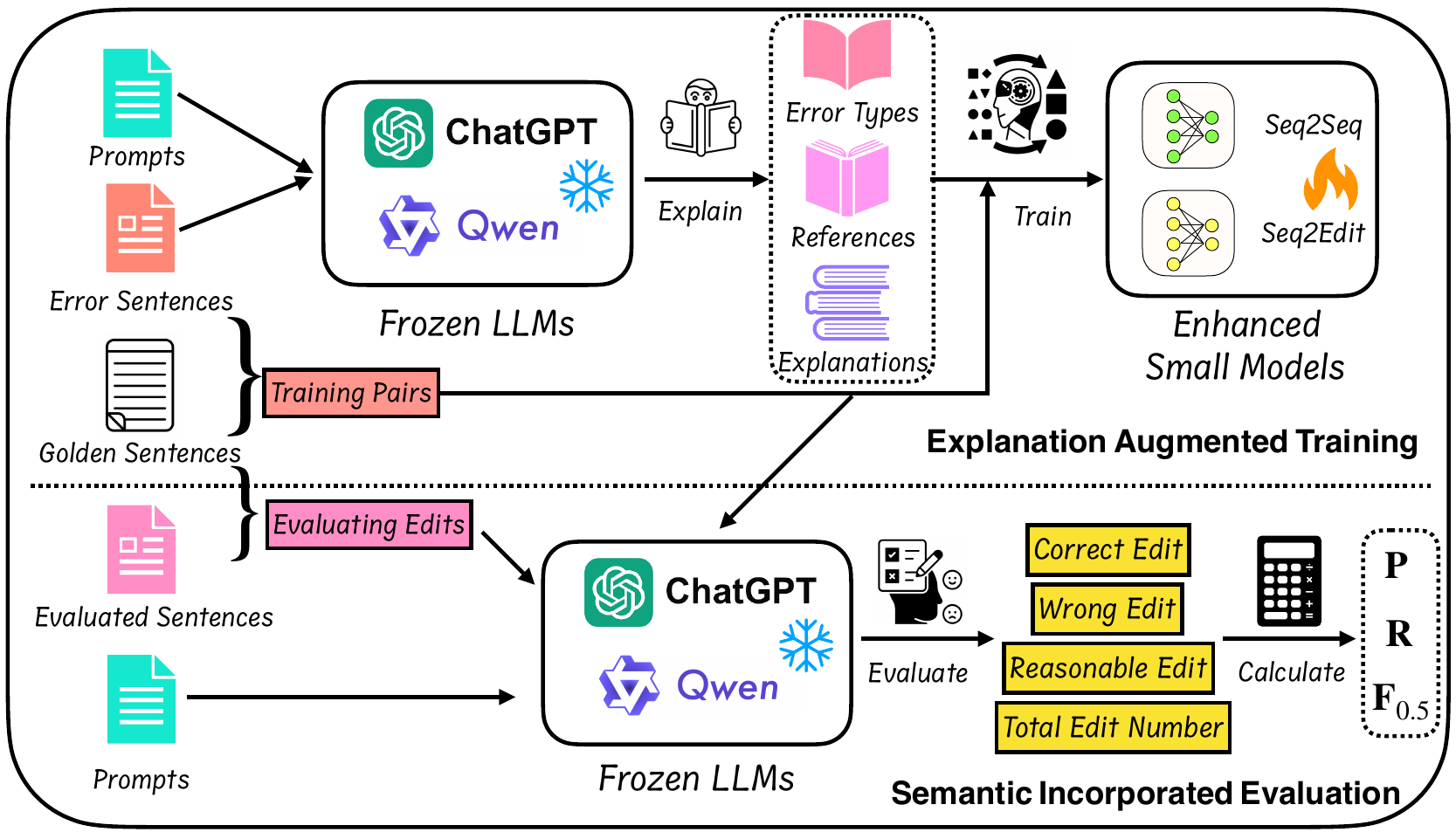}
\caption{Our designed frameworks of \ourmethod{} and \oureval{}.}
\label{Method_Figure}
\end{figure*}

\subsection{Explanation-Augmented Training}
As introduced in the above section, we propose the \textbf{EX}planation-\textbf{A}ug\textbf{M}ented training framework (\textbf{\ourmethod{}}) (as illustrated in Figure~\ref{Method_Figure}) to mine explanation information and grammatical knowledge from LLMs and inject them into small models, ultimately achieving the purpose of using LLMs to enhance the performance of small models. Based on our understanding of the CGEC task, we divide the explanation information (note that the ``explanation'' we consider here is the LLMs analysis of incorrect sentences in a broad sense) we want to obtain from LLMs into three categories:
\paragraph{Error Types} We believe that if the CGEC model knows the type of grammatical errors in the sentence to be corrected, it will help it reduce the search scope when correcting errors, thereby enabling it to make better corrections. Therefore, we ask LLMs to identify the error types based on the input sentences containing errors. Specifically, we pre-define types of common grammatical errors involving punctuation errors, spelling errors, word errors, syntax errors, etc. Then, we provide the defined error type schema along with the prompt to the LLMs, instructing them to choose only among the types we specified in the instruction prompt.

\paragraph{References} We observe that LLMs have a notable ability to generate correct sentences from incorrect ones, but the sentences they produce are not highly controllable. Although the sentences corrected by LLMs cannot be used as the final result, we believe they should serve as intermediate references for small models. Using corrections from LLMs as references can provide valuable cues to the small models, thereby enhancing their performance. Therefore, we also guide LLMs to make corrections they think are reasonable for the incorrect sentences and send the corrections provided by LLMs as references to the small model.

\paragraph{Explanations} To obtain high-quality explanations from LLMs, we define three dimensions of criteria to constrain LLMs: (1) \textit{Fluency} aims to ensure that the explanation text generated by LLMs has no grammatical errors and is fluent in expression; (2) \textit{Rationality} requires LLMs to explain grammatical errors as clearly and naturally as possible; (3) \textit{Comprehensiveness} is to ensure that all grammatical errors in the incorrect sentences can be explained as much as possible. Additionally, we also ask LLMs to rank multiple grammatical errors in a sentence according to error severity, that is, to generate explanations for important errors first.

After LLMs explain the samples in the dataset, we concatenate the obtained error types, references, and explanations to the front of the original input sentences. We then send the combined text to the small CGEC models for their training or inference. In summary, the design of \ourmethod{} is simple and intuitive. \textbf{LLMs and small models each perform their respective duties and give full play to their advantages.} The stored grammatical knowledge of LLMs is mined without additional fine-tuning. The small models take advantage of the alignment of supervised learning to downstream tasks with low training costs and obtain guidance from LLMs' task-related knowledge.

\subsection{Semantic-incorporated Evaluation}
To address the issue that traditional CGEC evaluation cannot flexibly adapt to the subjective nature of CGEC because they rely entirely on dataset annotation, we design the \textbf{SE}mantic-incorporated \textbf{E}valuation framework (\textbf{\oureval{}}). This framework utilizes LLMs to comprehensively evaluate CGEC by considering complex semantics.

Specifically, we first perform comparison and alignment preprocessing on the texts of error sentences and predicted sentences to obtain the predicted edits of the predicted text compared to the incorrect sentences.
We then require LLMs to evaluate each predicted edit from three dimensions based on grammatical analysis and semantic understanding of error sentences, golden sentences, and predicted sentences:
(1) \textit{Correct Edit} ($\mathbf{N}_{\text{CE}}$) indicates that LLMs judge the predicted edit to be effective in correcting the grammatical errors of the original sentence; 
(2) \textit{Wrong Edit} ($\mathbf{N}_{\text{WE}}$) signifies that LLMs determine that the predicted edit to be invalid and unable to correct grammatical errors; 
(3) \textit{Reasonable Edit} ($\mathbf{N}_{\text{RE}}$) refers to model edits not included in golden annotations, but which do not introduce new grammatical errors and do not affect the original semantics of the sentence. Usually, this type of edit involves some intonation particles and might be incorrectly classified as an incorrect edit by traditional metrics because it is not accounted for in the dataset annotations.
From these three dimensions we have designed, we can see that, \textbf{unlike different from traditional evaluation indicators, LLMs do not require precise text matching to determine whether the predicted edit exists in the golden edit set. Instead, the validity of the predicted edit is assessed more flexibly, taking into account the semantics of the text more comprehensively. }
In addition, it is worth mentioning that to make LLMs' judgment on edits more accurate, we also input the explanation information obtained in \ourmethod{} into LLMs at the same time when \oureval{} evaluates.

\begin{figure}[t]
\centering
\includegraphics[height=0.28\textwidth]{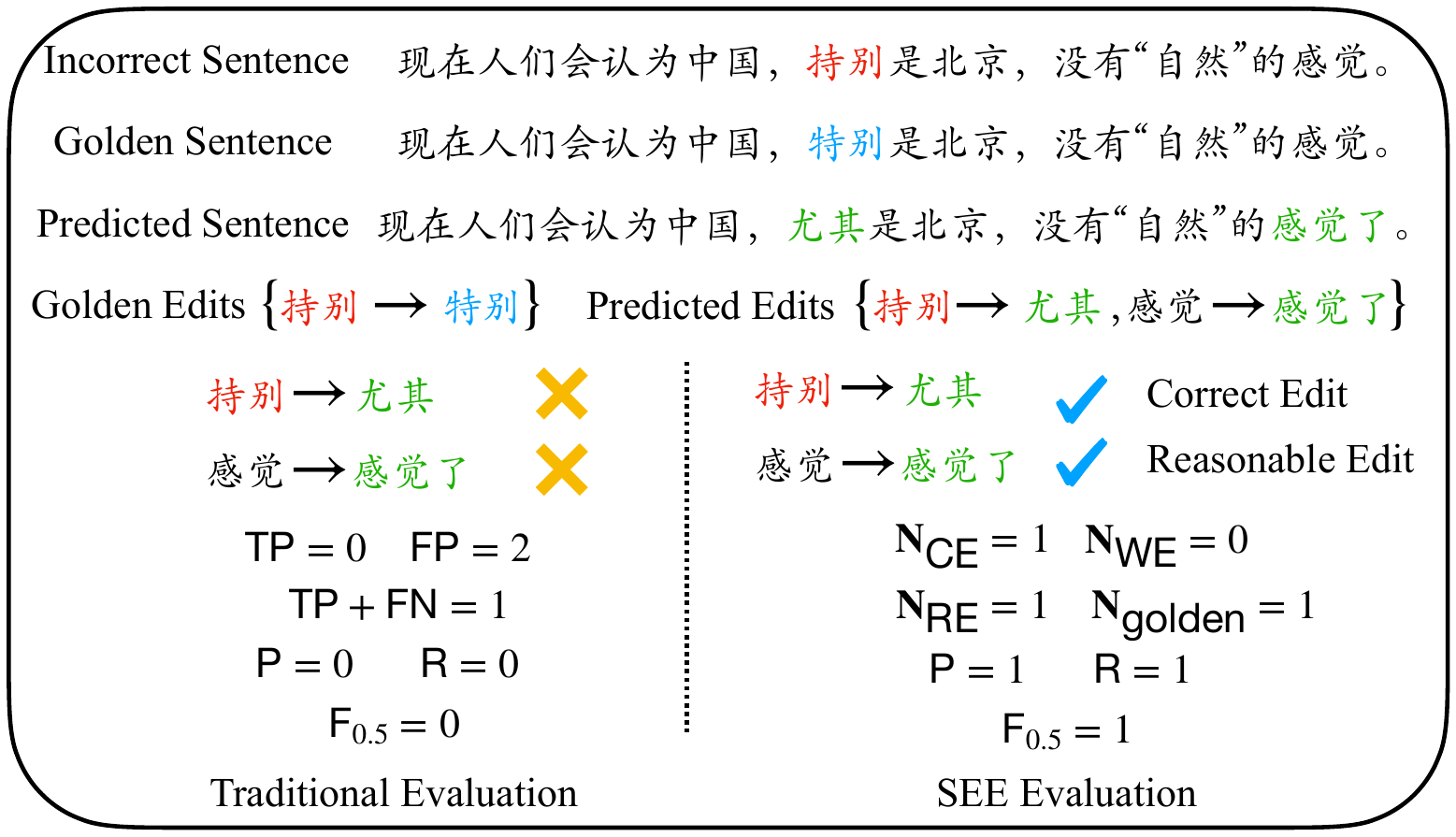}
\caption{The comparison examples of evaluation.}
\label{fig:eval}
\end{figure}

Based on the above three values derived from LLMs, we can calculate Precision, Recall, and $\text{F}_{0.5}$ scores as follows:

\begin{equation}\begin{aligned}
\text{P}=\frac{\mathbf{N}_{\text{CE}}}{\mathbf{N}_{\text{CE}} + \mathbf{N}_{\text{WE}}},
\end{aligned}\end{equation}

\begin{equation}\begin{aligned}
\text{R}=\frac{\mathbf{N}_{\text{CE}}}{\mathbf{N}_{\text{golden}}},
\end{aligned}\end{equation}

\begin{equation}\begin{aligned}
\text{F}_{0.5}=\frac{(1+0.5^2)\times \text{P} \times \text{R}}{0.5^2 \times \text{P} + \text{R}},
\end{aligned}\end{equation}
where $\mathbf{N}_{\text{golden}}$ is the length of the golden edit set for the incorrect sentence. The $\text{F}_{0.5}$ score is widely used in GEC-related studies because GEC is an application that pays more attention to precision. Furthermore, to better explain the mechanism of \oureval{}, we provide an evaluation example in Figure~\ref{fig:eval}.

To enable LLMs to perform the tasks we designed for \ourmethod{} and \oureval{}, we input both prompts and task demonstration examples into the LLMs to facilitate their adherence to our instructions through in-context learning. Due to the limitation of pages, the specific contents of our designed prompts for instructing LLMs to accomplish corresponding goals are presented in Appendix~\ref{sec:appendix_prompts}.

\section{Experiments}
\subsection{Experiment Setup}
\paragraph{Datasets} 
We mainly use the HSK dataset~\cite{zhang2009features} as training data. In our experiments, there are two settings for the use of training data: (1) \textbf{Full HSK data}, that is, using all 156,870 samples for model training; (2) \textbf{Sampled HSK data}, we randomly sample approximately 10\% of the HSK data, that is, 15,000 samples for model training.
In terms of test data, the CGEC data can be divided into two types of test data according to the source of the grammatical error sentences, namely Chinese-as-Second-Language (CSL) and Chinese native speaker data. To ensure the breadth of our experiment, we select the \textbf{NLPCC test data}~\cite{DBLP:conf/nlpcc/ZhaoJS018} which is the CSL data, and the \textbf{NaCGEC benchmark}~\cite{DBLP:conf/emnlp/MaLSZHZLLLCZS22} which is Chinese native speaker data as the test sets of our experiment. The NLPCC test data contains 2,000 samples and NaCGEC contains 5,869 incorrect sentences.

\paragraph{Evaluation Metrics}
To ensure the comparability of our experiments with previous CGEC works, in addition to using our own designed \textbf{\oureval{}} to evaluate P/R/$\text{F}_{0.5}$, we also report the widely used traditional \textbf{word/character-level} P/R/$\text{F}_{0.5}$. Particularly, as in the previous work~\cite{DBLP:conf/naacl/0004LBLZLHZ22}, we also apply the MaxMatch scorer~\cite{dahlmeier-ng-2012-better} and PKUNLP word segmentation tool~\cite{DBLP:conf/nlpcc/ZhaoJS018} to obtain the word-level performance. Therefore, to verify the effectiveness of our designed \ourmethod{}, we also conduct \textbf{human evaluation} experiments to provide the real performance of the models from a human perspective.

\paragraph{Baselines and Base Models}
The current mainstream CGEC models are mainly divided into two categories, namely Seq2Seq and Seq2Edit models. Since our \ourmethod{} framework is model-agnostic, we select the \textbf{representative Seq2Seq and Seq2Edit} models as baselines: 
(1) \textbf{BART-Large} \cite{katsumata-komachi-2020-stronger} and \textbf{mT5-Base}~\cite{DBLP:conf/naacl/XueCRKASBR21} are Seq2Seq models for text generation and can be straightforwardly trained for CGEC; 
(2) \textbf{GECToR-Chinese}~\cite{omelianchuk-etal-2020-gector} is the most widely used \textbf{Seq2Edit} method for CGEC.  
In addition, we select \texttt{GPT-3.5-Turbo}~\cite{DBLP:journals/corr/abs-2303-08774} and \texttt{Qwen-72B-Chat}~\cite{bai2023qwen} as the explainer-LLMs respectively. As for the evaluator-LLMs in \oureval{}, we recommend the most advanced \texttt{GPT-4-Turbo}~\cite{DBLP:journals/corr/abs-2303-08774}.

\paragraph{LLMs as Correctors} 
We selected two LLMs as Correctors to serve as baselines for comparison with our method. Specifically, we chose \texttt{Qwen-72B-Chat} and \texttt{GPT-3.5-Turbo} as our LLMs. We crafted a detailed prompt to ensure the LLMs deeply understood the task's significance when directly correcting Chinese grammatical errors (See Appendix~\ref{sec:appendix_prompt_LLM_Corrector}). Additionally, we experimented with in-context learning to enhance the performance of the LLMs. The experimental results and analysis of ``LLMs as Correctors'' are presented in Appendix~\ref{sec:appendix_results_LLM_Corrector}.

\begin{table*}[htb]
\small
\centering
\begin{tabular}{@{}l|l|ccc|ccc|ccc@{}}
\toprule
 \multicolumn{1}{c|}{\textbf{Training Data}} & \multicolumn{1}{c|}{\textbf{Model}} & \multicolumn{3}{c|}{\textbf{Word-Level}}          & \multicolumn{3}{c|}{\textbf{Character-Level}}   & \multicolumn{3}{c}{\textbf{\oureval{}}}        \\
  &  & \textbf{P}  & \textbf{R}  & \textbf{$\textbf{F}_{0.5}$}  & \textbf{P}  & \textbf{R} & \textbf{$\textbf{F}_{0.5}$} & \textbf{P}  & \textbf{R} & \textbf{$\textbf{F}_{0.5}$}           \\ 
\midrule
 None  & GPT-3.5-Turbo              & 24.36          & 28.01          & 25.01          & 27.71          & 29.19          & 27.99 & 53.82 & 30.14	& 46.51       \\
 None  & Qwen-72B-Chat              & \textbf{27.88} &	\textbf{32.85} &	\textbf{28.75}	& \textbf{32.42} &	\textbf{34.97} &	\textbf{32.90} &	\textbf{67.20} & \textbf{35.01} & \textbf{56.76}    \\
 \midrule  
Sampled (15K)  & mT5-Base              & 16.10 & 8.93 &	13.87 & 30.25 &	8.77 & 20.30  & 58.36 &	9.89 &	29.47          \\
Full (156K) & mT5-Base          & 24.08 & 16.74 & 22.14 & 38.37 &	17.14 &	30.75 & 67.37 &	19.37 &	45.05  \\
Sampled (15K) &  w/ \ourmethod{} (GPT)        & 25.21$^\uparrow$ & 17.76$^\uparrow$ &	23.26$^\uparrow$	& \textbf{39.04}$^\uparrow$ &	18.16$^\uparrow$ &	31.74$^\uparrow$ &	69.29$^\uparrow$ &	20.27$^\uparrow$ &	46.70$^\uparrow$ \\
Sampled (15K) &  w/ \ourmethod{} (Qwen)      & \textbf{26.41}$^\uparrow$ &	\textbf{20.57}$^\uparrow$ &	\textbf{25.00}$^\uparrow$ &	38.76$^\uparrow$ &	\textbf{21.81}$^\uparrow$ &	\textbf{33.55}$^\uparrow$ &	\textbf{69.76}$^\uparrow$ &	\textbf{22.63}$^\uparrow$ &	\textbf{49.25}$^\uparrow$  \\
                                        \midrule 
 Sampled (15K)  & BART-Large          & 19.46 &	14.77 &	18.30 &	32.07 &	13.67 &	25.27 &	62.94 &	12.18 &	34.33 \\
 Full (156K) & BART-Large        & \textbf{28.35} &	22.30 & 	26.89 &	39.10 &	22.75 &	34.19 & 63.16 &	17.31 &	41.29          \\ 
 Sampled (15K)  &  w/ \ourmethod{} (GPT)            & 28.33$^\uparrow$ &	\textbf{23.38}$^\uparrow$ &	\textbf{27.17}$^\uparrow$ &	39.61$^\uparrow$ &	\textbf{23.87}$^\uparrow$ &	\textbf{35.00}$^\uparrow$	& \textbf{68.55}$^\uparrow$ & \textbf{23.31}$^\uparrow$ & \textbf{49.38}$^\uparrow$         \\
 Sampled (15K)  &  w/ \ourmethod{} (Qwen)              & 27.91$^\uparrow$ &	22.24$^\uparrow$ &	26.55$^\uparrow$ &	\textbf{40.01}$^\uparrow$ &	22.90$^\uparrow$ &	34.81$^\uparrow$ &	62.94$^\uparrow$	& 22.18$^\uparrow$ &	46.02$^\uparrow$    \\
                                        
\midrule
Sampled (15K) & GECToR-Chinese              & 10.85 &	6.40 &	9.53 &	\textbf{29.49} &	4.65 &	14.26 & 55.60 &	4.41 &	16.74       \\
Full (156K) & GECToR-Chinese          & \textbf{18.26} &	10.99 &	16.12 &	27.03 &	11.99 &	21.60 &	48.32 &	\textbf{12.21} &	30.36  \\
Sampled (15K) &  w/ \ourmethod{} (GPT)         & 18.09$^\uparrow$ &	\textbf{12.74}$^\uparrow$ &	\textbf{16.69}$^\uparrow$ &	27.53$^\uparrow$ &	\textbf{12.71}$^\uparrow$ &	\textbf{22.32}$^\uparrow$ &	\textbf{49.46}$^\uparrow$ &	12.05$^\uparrow$ &	\textbf{30.51}$^\uparrow$  \\
Sampled (15K) &  w/ \ourmethod{} (Qwen)         & 17.31$^\uparrow$ &	12.06$^\uparrow$ &	15.92$^\uparrow$ &	25.95$^\uparrow$ &	11.63$^\uparrow$ &	20.82$^\uparrow$ &	48.98$^\uparrow$	& 11.49$^\uparrow$ & 29.63$^\uparrow$ \\                         
\bottomrule
\end{tabular}
\caption{
Performance of various models on the NLPCC test set. Note that 15K and 156K represent the amount of HSK data. $^\uparrow$ means that \ourmethod{} has improved performance compared to the baselines with the same training data.}
\label{Table: Main_Results_NLPCC}
\end{table*}

\paragraph{Implementation Details} 
We utilize Chinese-BART-Large \cite{DBLP:journals/corr/abs-2109-05729}, Mengzi-T5-Base (Chinese)~\cite{DBLP:journals/corr/abs-2110-06696}, Chinese-Struct-Bert-Large~\cite{DBLP:conf/iclr/0225BYWXBPS20} to initialize small models.
For open-source LLMs, we run their inference process on 4 NVIDIA A100 GPUs. For closed-source LLMs, we directly access them through the official APIs.
It is worth noting that in all our reported experiments, \ourmethod{} provides only one error type/reference/explanation information for each incorrect sentence. Because our experiments are only verification experiments, for better performance, researchers can obtain more explanation information to enhance the small models in \ourmethod{}.
The specific prompts used by our method are in Appendix~\ref{sec:appendix_prompts}, and other implementation details and hyperparameter selection are in Appendix~\ref{sec:appendix_hyper}.

\begin{table}[t]
\small
\centering

\begin{tabular}{@{}l|cc@{}}
\toprule
\multicolumn{1}{c|}{\textbf{Method}}  & \textbf{Word-$\text{F}_{0.5}$} & \textbf{Char-$\text{F}_{0.5}$}\\
\midrule
~~BART-Large  & 18.30 & 25.27   \\
\midrule

 + Error Types  & 21.74$^\uparrow$ & 29.12$^\uparrow$  \\
 + References  & 23.88$^\uparrow$ & 33.49$^\uparrow$  \\
 + Explanations  & 21.52$^\uparrow$ & 29.84  \\
 + Error Types~ + References & 24.21$^\uparrow$ & 33.66$^\uparrow$  \\ 
 + Error Types~ + Explanations & 23.29$^\uparrow$ & 32.54$^\uparrow$  \\
 + References~ + Explanations & 25.18$^\uparrow$ & 33.74$^\uparrow$  \\
\midrule
~~BART-Large w/ \ourmethod{} (GPT)  & \textbf{27.17} & \textbf{35.00}   \\

\bottomrule
\end{tabular}

\caption{Ablation results for fine-grained explanation information. The training data for all models is 15K sampled HSK data. The test data is NLPCC. Note that the BART-Large w/ \ourmethod{}{} (GPT) is equivalent to BART-Large+Error Types+References+Explanations.}
\label{tab:ablation_explanation}
\end{table}

\subsection{Main Results}
Our main results on NLPCC are presented in Table~\ref{Table: Main_Results_NLPCC}, we also provide main results and analyses on NaCGEC in Appendix~\ref{sec:appendix_main_results_nacgec} and Table~\ref{Table: Main_Results_NaCGEC}.

\paragraph{Main Results of \ourmethod{}}
From Table~\ref{Table: Main_Results_NLPCC}, we can know that: (1) With the same amount of training data, \ourmethod{} generally brings significant improvements to all baselines under all evaluation metrics. (2) With only 10\% of the labeled training data, small models enhanced by \ourmethod{} achieve performance equivalent to or better than that of training with the full amount of data. (3) The model-agnostic nature of \ourmethod{} enables it to bring stable gains no matter what LLMs are selected, or for small models of Large/Base scale.

\paragraph{Main Results of \oureval{}}
From Table~\ref{Table: Main_Results_NLPCC}, we see that: (1) The evaluation results of \oureval{} are basically consistent in trend with traditional metrics, which shows the correctness of \oureval{}. (2) Especially for the results of LLMs, we observe that \oureval{} achieves a huge numerical difference from the results obtained by traditional metrics, which indicates that \oureval{} is more suitable for GEC evaluation in the era of LLMs. Note that the base model of \oureval{} is \texttt{GPT-4-Turbo}, which is different from the evaluated LLMs, so it will not cause unfair evaluation. 
Moreover, we propose the hypothesis that our \oureval{} evaluation metric aligns more closely with human judgment, which will be discussed in Section~\ref{sec:human_eval}. Therefore, small models trained with \ourmethod{} can exhibit competitive performance with LLMs. This could be more meaningful in real-world scenarios, as deploying a small model is more cost-effective and offers faster response times.

\subsection{Analyses and Discussion}
\subsubsection{The Impact of Fine-grained Explanation Information on \ourmethod{}}
The main results of \ourmethod{} are derived from three kinds of information error types/references/explanations from LLMs. Therefore, it is necessary to conduct ablation studies on the three kinds of information to assess their respective contributions to \ourmethod{}.
As shown in Table~\ref{tab:ablation_explanation}, we conduct ablation experiments on NLPCC test data with \texttt{GPT-3.5-Turbo} as the base model of \ourmethod{} and BART-Large as the enhanced small model. 
We can see that each type of information can bring significant improvements to BART-Large when executed individually, demonstrating the correctness of our choice of obtaining information from LLMs.
In particular, the references have the greatest improvement for the small model, which shows that the correction results made by LLMs can bring good reference and guidance to the small model, and a good reference correction result can bring the most direct gain to the small model.
Furthermore, we see that when various types of information are used in pairs, performance can be further improved compared to individual information. This shows that the compatibility between the three types of information we designed is very good and would not affect each other.

\begin{table}[t]
\small
\centering

\begin{tabular}{@{}l|cc@{}}
\toprule
\multicolumn{1}{c|}{\textbf{Method}}  & \textbf{Word-$\text{F}_{0.5}$} & \textbf{Char-$\text{F}_{0.5}$}\\
\midrule
~~BART-Large  & 18.30 & 25.27   \\
\midrule
Train (No gold) / Test (No gold) & 27.17$^-$ & 35.00$^-$  \\
Train (Gold) / Test (No gold) & 21.57$^\downarrow$ & 28.93$^\downarrow$  \\ 
Train (No gold) / Test (Gold) & 25.98$^\downarrow$ & 37.56$^\uparrow$  \\
Train (Gold) / Test (Gold) & 43.10$^\uparrow$ & 60.40$^\uparrow$  \\
\midrule
~~BART-Large w/ \ourmethod{} (GPT)  & 27.17 & 35.00  \\

\bottomrule
\end{tabular}

\caption{The impact of golden annotation information. The training data is 15K sampled HSK data. The test data is NLPCC.  Note that the BART-Large w/ \ourmethod{}{} (GPT) is equivalent to Train (No gold) / Test (No gold).}
\label{tab:ablation_gold}
\end{table}

\subsubsection{The Impact of Golden Annotation Information on \ourmethod{}}
To further explore the performance upper bound of \ourmethod{}, in the process of using LLMs to obtain training and test data for the small model, we input the golden sentences annotated by the dataset into the LLMs to observe the performance changes of the small model. In other words, we want to observe how the quality of the explanation information generated by LLMs changes when they are provided with golden sentences as input.
In Table~\ref{tab:ablation_gold}, we are surprised to find that when we add golden sentences in the process of LLMs generating training data or generating test data, the model performance declines compared to not adding golden sentences in both processes (i.e., Train (No gold)/ Test (No gold)). This is an interesting and counter-intuitive phenomenon, and we believe it highlights the difference and gap between the generative paradigm of LLMs and the golden sentences annotated in the dataset. If LLMs are only allowed to see golden sentences during training or testing, 
the explanation information they generate will differ significantly from what they would typically produce on their own. This discrepancy can create a gap between the training and test data of the small model, leading to performance degradation.
Therefore, we can also understand why there is a huge performance gain when inputting golden sentences to LLMs in both training and testing processes. In this case, LLMs generate sentences similar to golden sentences in both training data and test data.
\begin{figure}[htb]
\centering
\includegraphics[height=0.29\textwidth]{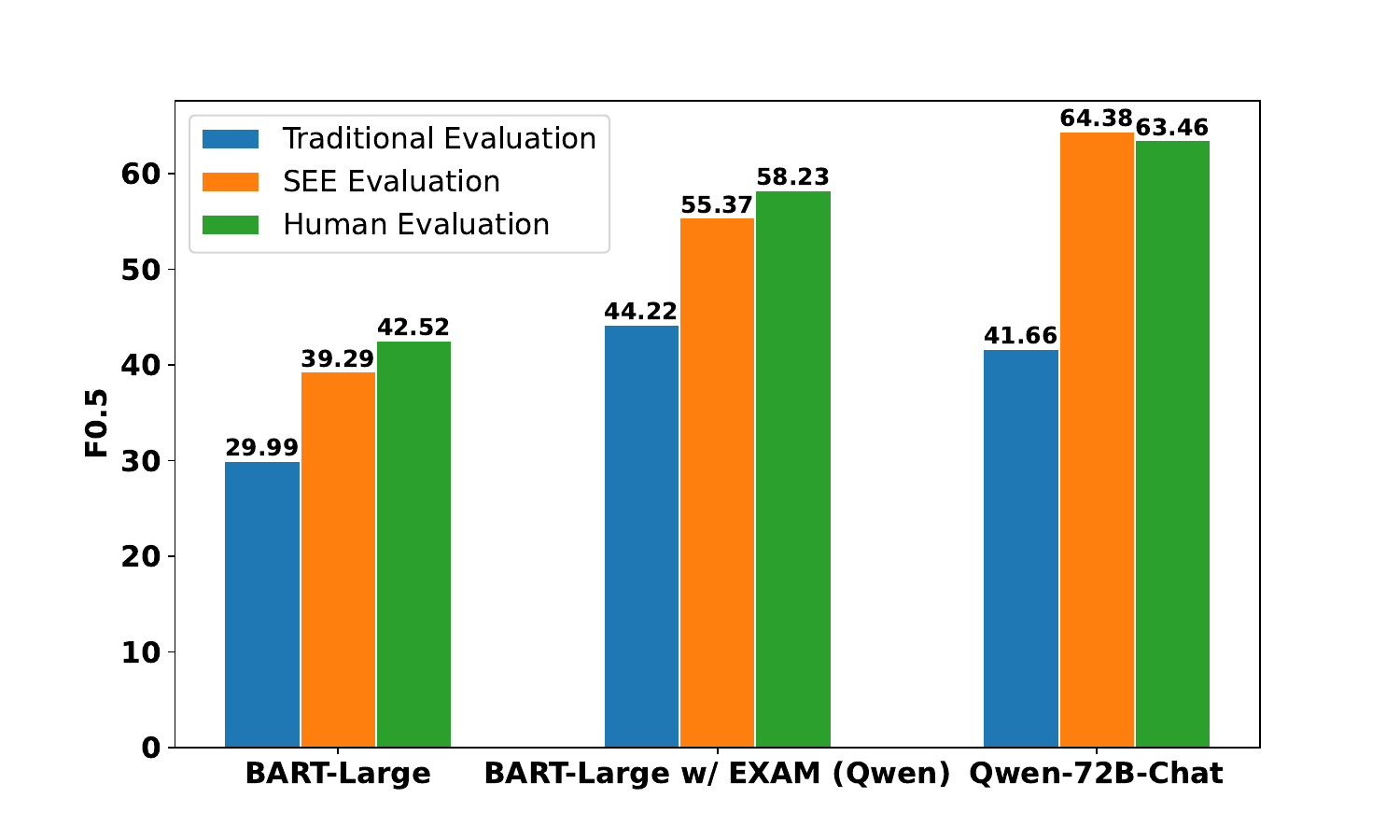}
\caption{Human evaluation results. The training data is 15K sampled HSK data. The test data is 200 sampled NLPCC data. The traditional metric is Char-$\text{F}_{0.5}$.}
\label{fig:human_eval}
\end{figure}

\subsubsection{Human Evaluation for \oureval{}}
\label{sec:human_eval}
The design motivation of \oureval{} is to use LLMs to bring evaluation more consistent with the human perspective to CGEC. Therefore, we conduct human evaluation experiments to observe whether \oureval{} or traditional metrics are closer to human. Specifically, we randomly select 200 test samples from NLPCC, then have three annotators to independently evaluate the models' correct results. We calculate the average P/R/$\text{F}_{0.5}$ scores of human evaluation based on the judegments from the three annotators. From Figure~\ref{fig:human_eval}, we see that: (1) For various models, \oureval{}'s evaluation is closer to human evaluation than traditional evaluation, which shows that our designed \oureval{} can more realistically measure the CGEC performance than traditional evaluation. (2) \oureval{}'s evaluation of LLMs differs very little from human evaluation, indicating that \oureval{} is more suitable for the evaluation of LLMs. (3) Unlike the cases where evaluation results for small models fall below human evaluation, \oureval{}'s evaluation of LLMs can slightly surpasses human evaluation results. This is because \oureval{} relies on another LLM (i.e., \texttt{GPT-4-Turbo}) for its evaluation process, indicating better understanding among LLMs.

\begin{table*}
  \small
  \centering
  \begin{tabular}[htb]{cl}
    \toprule
    \textbf{Error Sentence} & \zh{这段话给我们有道理的，虽然现在黑暗，但等着、忍着，光明会到来的。 } \\
    \midrule
    \multirow{2}{*}{\textbf{Golden Sentence}} & \zh{这段话是有道理的，虽然现在黑暗，但等着、忍着，光明就会到来。} \\
    & This paragraph is reasonable. Although it is dark now, if we wait and endure, the light will come. \\
    \midrule
    \textbf{Error Type} & \zh{标点误用，句式杂糅} \\
    \textbf{\texttt{GPT-3.5-Turbo}} & Misuse of punctuation, mixed sentence structures \\
    \midrule
    \textbf{Reference} & \multirow{2}{*}{\zh{这段话给我们\textcolor{orange}{的道理：}虽然现在黑暗，但等着、忍着，光明会到来的。}} \\
    \textbf{\texttt{GPT-3.5-Turbo}} &  \\
    \midrule
    \textbf{Explanation} & \zh{在这个句子中，动词“给我们”和“有道理”\textcolor{red}{成分赘余}，此处改为“给我们的道理”，} \\
    \textbf{\texttt{GPT-3.5-Turbo}} & \zh{并且下句为道理的内容，因此不应该使用“，”而应该使用表示承接的“：”。} \\
    \midrule
    \textbf{Error Type} & \zh{词语冗余, 逻辑不通} \\
    \textbf{\texttt{Qwen-72B-Chat}} & Redundant words, illogical logic \\
    \midrule
    \textbf{Reference} & \zh{这段话给我们\textcolor{orange}{提供了有道理的启示}，虽然现在\textcolor{orange}{处于黑暗时期}，但\textcolor{orange}{只要坚持等待}、}\\
    \textbf{\texttt{Qwen-72B-Chat}} & \zh{\textcolor{orange}{忍受}，光明\textcolor{orange}{总}会到来的。} \\
    \midrule
    \textbf{Explanation} & \zh{“给我们”与“有道理的”都表示“这段话”的意义，造成了词语冗余。应将“给我们”改} \\ \multirow{2}{*}{\textbf{\texttt{Qwen-72B-Chat}}} & \zh{为“提供了”。接着，“虽然现在黑暗，但等着、忍着，光明会到来的”缺乏逻辑连贯} \\
     &  \zh{性，\textcolor{red}{应补充关联词语“虽然”。}} \\
    \bottomrule
  \end{tabular}
  \caption{We mark changes made by LLMs in \textcolor{orange}{orange} and poor explanations given by LLMs in \textcolor{red}{red}.}
  \label{tab: case_study}
\end{table*}

\subsection{Case Observation}
\label{exp:case}
To verify the correctness of our motivation for using LLMs as explainers, and to demonstrate the explanation information generated by \ourmethod{}, we give cases in Table~\ref{tab: case_study} of \texttt{GPT-3.5-Turbo} and \texttt{Qwen-72B-Chat} acting as the explainer respectively. As seen in Table~\ref{tab: case_study}, although the two LLMs make different error-type judgments, both provide reasonable explanations for their respective judgments. 
Regarding the reference corrections they give, we see that \texttt{Qwen-72B-Chat} prefers free generation compared to \texttt{GPT-3.5-Turbo}. Of course, we think the corrected sentence generated by \texttt{Qwen-72B-Chat} is more fluent and reasonable. 
For the explanations of grammatical errors made in the incorrect sentence, we can see that both LLMs give quality explanations to a certain extent. Although there are some minor flaws, overall, they can give explanations that can be helpful for humans or small models to improve. 
Additionally, we include more cases where LLMs offer explanations and evaluations in the form of data supplementary materials.
\section{Conclusion}
In this paper, focusing on the dilemma that LLMs cannot achieve satisfactory results as correctors on CGEC, we rethink how LLMs should be effectively utilized in the CGEC task. 
To fully exploit the rich grammatical knowledge and powerful semantic understanding ability of LLMs, and bypass the main reason why the LLMs corrector is not suitable for the CGEC task, that is, the minimum change principle, we propose the training framework \ourmethod{} that uses LLMs as explainers to enhance CGEC small models, and the novel evaluation method \oureval{} that utilizes LLMs as evaluators to give more reasonable evaluation of the CGEC task.
Extensive empirical results and analyses show that our work is a meaningful exploration of how LLMs and small models can coexist and make progress together on downstream tasks such as CGEC.

\section*{Limitations}
Currently, the main limitation of our work is the scope of the languages. As we all know, GEC in various languages has its application significance, so it is valuable to apply our methods to other languages further. 
The main reason why we did not apply our methods to languages such as English is that there are many differences in the types of grammatical errors and grammatical rules that CGEC and EGEC focus on. 
Therefore, the prompts of \ourmethod{} and \oureval{} need to be re-customized when applied to the English scenario. 
The purpose of our paper is to rethink how LLMs should be appropriately utilized in the GEC field. 
Changing prompts to adapt to new languages is not the main technical contribution and innovation we pursue.
In the future, to enhance the impact of our work and serve a wider community, we will expand \ourmethod{} and \oureval{} to the English scenario.

\section*{Ethics Statement}
Our used data and models (including LLMs) are all publicly available academic resources. We also paid for closed-source LLMs that require charging for APIs, so there is no ethical issue about data or models in our work.

\section*{Acknowledgments}
This research is supported by National Natural Science Foundation of China (Grant No. 62276154), Research  Center for Computer Network (Shenzhen) Ministry of Education, the Natural Science Foundation of Guangdong Province (Grant No. 2023A1515012914), Shenzhen Science and Technology Program (Grant No. WDZC2023112809143 7002), Basic Research Fund of Shenzhen City (Grant No. JCYJ20210324120012033 and GJHZ202402183000101), the Major Key Project of PCL for Experiments and Applications (PCL2021A06).

\section*{Declaration of interest}
The authors declare that they have no known competing financial interests or personal relationships that could have appeared to influence the work reported in this paper.

\section*{Data availability}
The data used in our work will be made available on request.

\printcredits

\bibliographystyle{model1-num-names}

\bibliography{cas-refs}

\clearpage
\appendix
\section{Prompt of LLMs as Corrector}
\label{sec:appendix_prompt_LLM_Corrector}
To enable the LLM to directly provide corrected versions of the original sentences, we used the following prompt:

\zh{请你针对给出的中文文本中的标点错误、拼写错误、词语错误和句法错误等提供合理且忠实的纠正。

例如：

<SOURCE SENTENCE> 纠正为：<TARGET SENTENCE>

请你纠正（直接输出纠正后的句子，无需任何解释）：

<SOURCE SENTENCE>
}

\section{Results and Analysis of LLMs as Corrector}
\label{sec:appendix_results_LLM_Corrector}
\begin{figure}[t]
\centering
\includegraphics[height=0.25\textwidth]{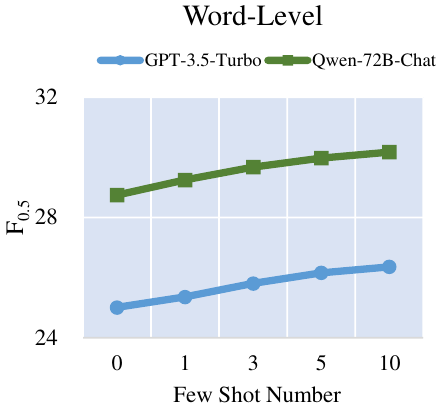}
\caption{Few-shot results of LLMs on the word-level metric.}
\label{fig:w-level}
\end{figure}

\paragraph{Results}
In Table~\ref{Table: Main_Results_NLPCC}, we observe that in the zero-shot scenario, GPT-3.5-Turbo scores 25.01 and 27.99 for Word-Level and Character-Level $\text{F}_{0.5}$, respectively, while \texttt{Qwen-72B-} \texttt{Chat} scores 28.75 and 32.90. Under our proposed \oureval{} evaluation method, the $\text{F}_{0.5}$ scores for \texttt{GPT-3.5-Turbo} and \texttt{Qwen-72B-Chat} are 46.51 and 56.76, respectively. Figure~\ref{fig:w-level} and Figure~\ref{fig:w-level} show that in the few-shot scenario, both \texttt{GPT-3.5-Turbo} and \texttt{Qwen-72B-Chat} improve their scores at the Word-Level and Character-Level.
\paragraph{Analysis}
From these experimental results, it is evident that even with the enhancement provided by few-shot learning, there remains a significant gap in the correction capabilities of LLMs. Despite their strong language generation abilities, current LLMs score lower than smaller models under traditional evaluation metrics, which does not align with human perception, as seen in Figure~\ref{fig:human_eval}. However, our SEE method maintains a high level of alignment with human judgment.

\begin{figure}[t]
\centering
\includegraphics[height=0.25\textwidth]{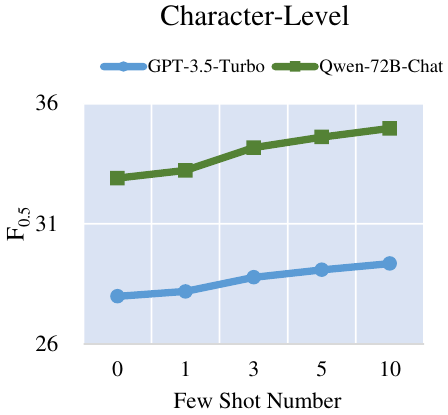}
\caption{Few-shot results of LLMs on the character-level metric.}
\label{fig:c-level}
\end{figure}

\section{Our Designed Prompts for \ourmethod{} and \oureval{}}
\label{sec:appendix_prompts}
\begin{figure*}[h]
\centering
\includegraphics[width=0.78\textwidth]{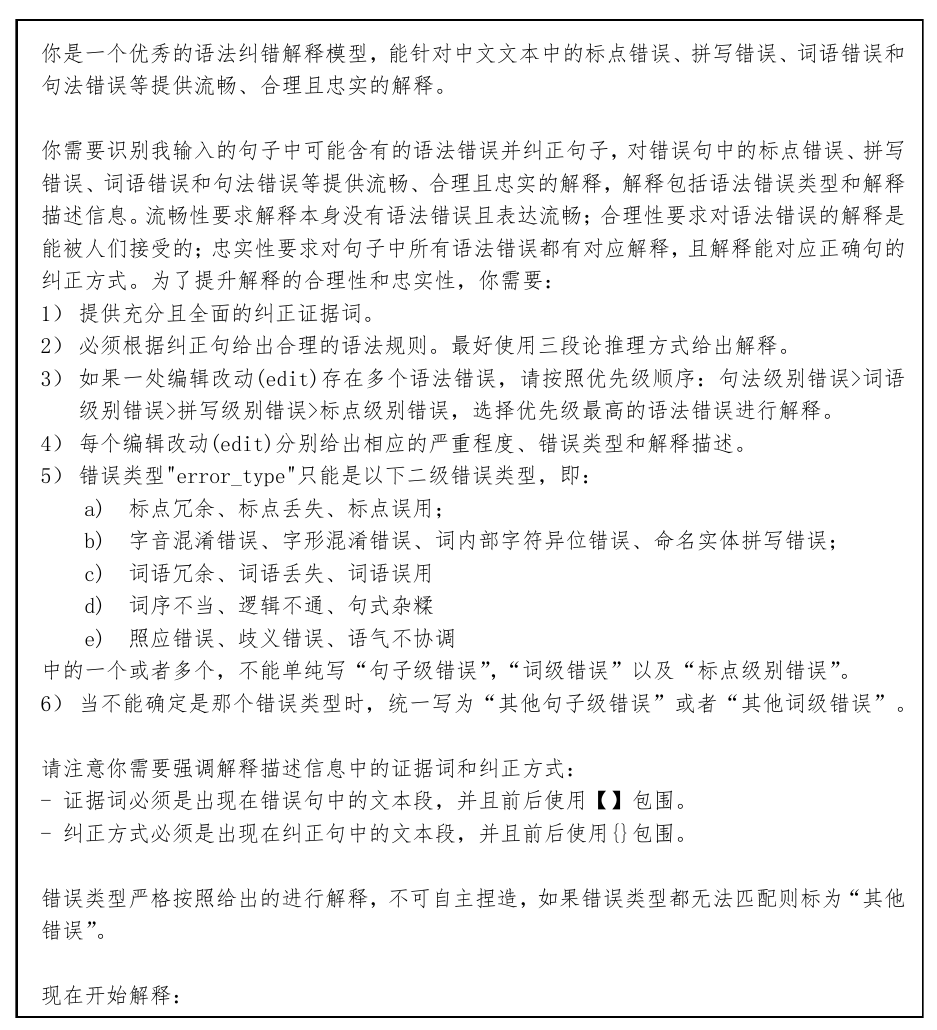}
\caption{Our designed explanation prompt for \ourmethod{}.}
\label{Explain_Prompt_Figure}
\end{figure*}

\begin{figure*}[h]
\centering
\includegraphics[width=0.78\textwidth]{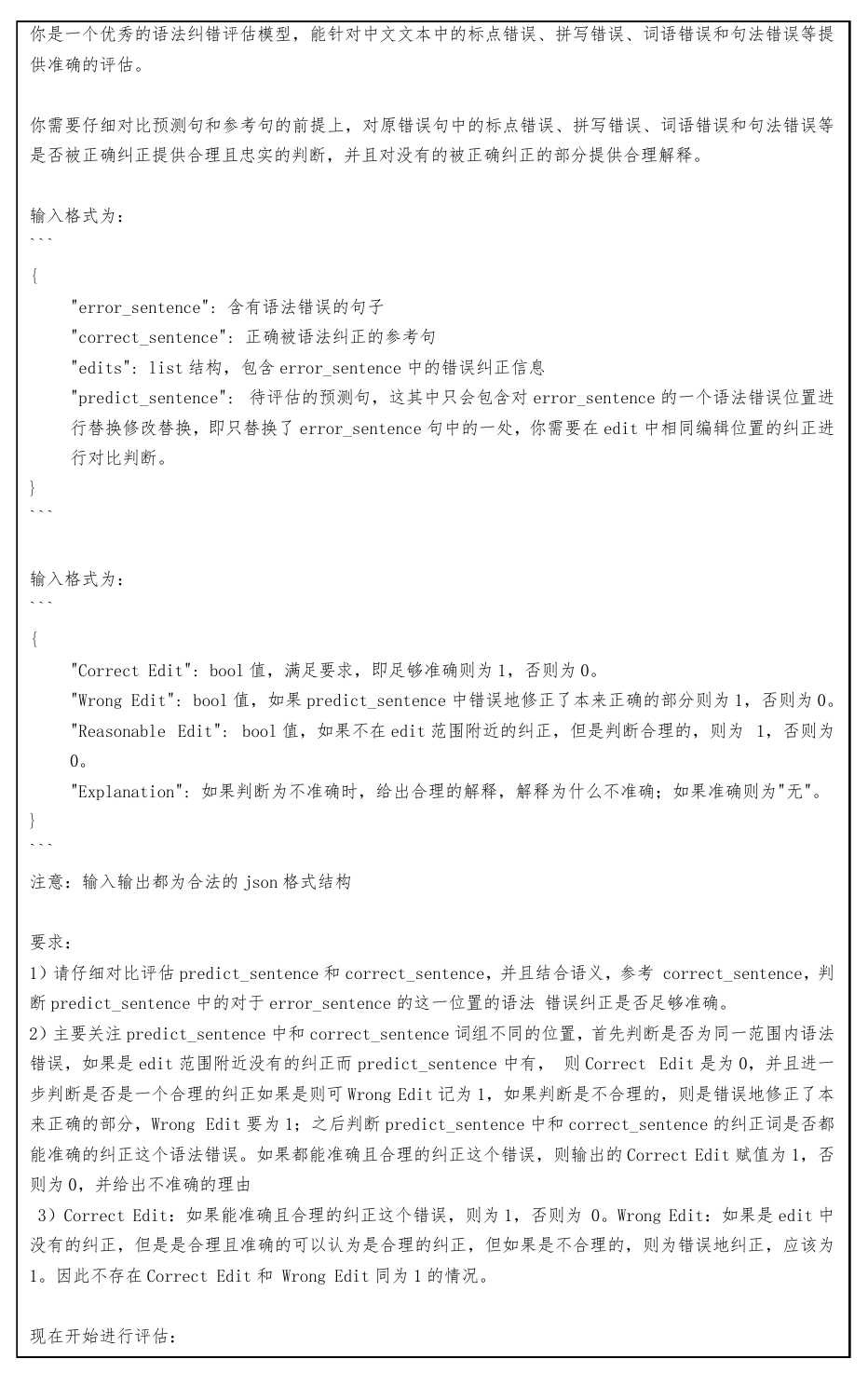}
\caption{Our designed evaluation prompt of \oureval{}.}
\label{Evaluate_Prompt_Figure}
\end{figure*}

In order to guide LLMs to achieve our designed tasks as we expect, we carefully design the instruction prompts based on the characteristics of the CGEC task. The prompts for explanation are as shown in Figure~\ref{Explain_Prompt_Figure}, and the prompts for evaluation are as shown in Figure~\ref{Evaluate_Prompt_Figure}. In addition, as mentioned in the main text of this paper, to make the results generated by LLMs more accurate, we also input task examples (or demonstrations) to LLMs to stimulate their In-context Learning capabilities.
Considering that the prompts with in-context learning examples added are very long, we upload the prompts with task examples in the form of software supplementary materials to facilitate peer review.

\section{Implementation Details and Hyperparameters}
\label{sec:appendix_hyper}
The hyperparameter values of the small models in our experiments are shown in Table~\ref{tab:hyper}. Besides, the loss functions for Seq2Seq models are the label-smoothed cross-entropy, and the loss function for Seq2Edit is cross-entropy.

\begin{center}
\begin{table*}[htbp]
\centering
\begin{tabular}{cccc}
\toprule
\textbf{Configurations}   & \textbf{BART-Large}  & \textbf{mT5-Base} &  \textbf{GECToR-Chinese}                              \\ \hline
Model type        & Seq2Seq & Seq2Seq & Seq2Edit                               \\
Epochs          & 10   &  10 & 20 (2 cold epochs)      \\                       
Batch size        & 256 & 256 & 128                                              \\
Optimizer                 &  Adam  &  Adam &  Adam \\ 
$\beta_1$ & 0.9 & 0.9 & 0.9 \\
$\beta_2$ & 0.999 & 0.999 & 0.999 \\
$\epsilon$ & $ 1 \times 10^{-8}$  & $ 1 \times 10^{-8}$ & $ 1 \times 10^{-8}$     \\
Learning rate             & $3 \times 10^{-6}$     & $5 \times 10^{-5}$  & $1 \times 10^{-5} (1 \times 10^{-3} \text{for cold})$                                     \\
\bottomrule
\end{tabular}
\caption{Hyperparameter values of the small models to be enhanced in our experiments.}
\label{tab:hyper}
\end{table*}
\end{center}

\section{Main Results on NaCGEC}
\label{sec:appendix_main_results_nacgec}
The main results of \ourmethod{} and \oureval{} on NaCGEC are presented in Table~\ref{Table: Main_Results_NaCGEC}. Note that the models we test on NaCGEC are all trained using HSK data. The HSK data comes from sentences with grammatical errors made by foreigners when learning Chinese, while NaCGEC comes from the grammatical errors made by native Chinese speakers in daily life. 
\citeauthor{DBLP:conf/emnlp/MaLSZHZLLLCZS22} have proven that Chinese native CGEC data such as NaCGEC is more difficult than CSL data such as HSK because the grammatical errors made by native speakers are more subtle than those made by foreigners. 
Therefore, as shown in Table~\ref{Table: Main_Results_NaCGEC}, when CGEC models trained with HSK data are tested on NaCGEC, low performance is understandable and expected.

From Table~\ref{Table: Main_Results_NaCGEC}, we can get similar conclusions as on NLPCC. \ourmethod{} can bring stable and competitive enhancements to small models with the participation of small-scale training data, and the performance enhanced by \ourmethod{} is comparable to the performance of small models trained with full-scale data. Meanwhile, \oureval{} can still bring reliable evaluation to CGEC models. The experiment on NaCGEC reflects the robustness of our proposed \ourmethod{} and \oureval{} to different data sources, that is, they are effective for both CSL CGEC data and native CGEC data.

\begin{table*}[ht]
\small
\centering
\begin{tabular}{@{}l|l|ccc|ccc|ccc@{}}
\toprule
 \multicolumn{1}{c|}{\textbf{Training Data}} & \multicolumn{1}{c|}{\textbf{Model}} & \multicolumn{3}{c|}{\textbf{Word-Level}}          & \multicolumn{3}{c|}{\textbf{Character-Level}}   & \multicolumn{3}{c}{\textbf{\oureval{}}}        \\
  &  & \textbf{P}  & \textbf{R}  & \textbf{$\textbf{F}_{0.5}$}  & \textbf{P}  & \textbf{R} & \textbf{$\textbf{F}_{0.5}$} & \textbf{P}  & \textbf{R} & \textbf{$\textbf{F}_{0.5}$}           \\ 
\midrule
 None  & \texttt{GPT-3.5-Turbo}              & 13.84 & \textbf{11.67}	& 13.35 & 9.58 & \textbf{9.66} & 9.59  & \textbf{39.65} & \textbf{12.17} & \textbf{27.31}z    \\
 None  & \texttt{Qwen-72B-Chat}              & \textbf{14.23} & 11.33 & \textbf{13.53} & \textbf{10.32}	& 8.83& \textbf{9.98} & 32.55 &	4.74 & 23.14          \\
 \midrule  
Sampled (15K)  & mT5-Base              & 5.38 &	0.65 & 2.19 & 4.5 &	0.64 & 2.03 & \textbf{36.11} &	4.40 &	14.79     \\
Full (156K) & mT5-Base          & 2.78 & 3.72 &	2.93 & 1.98 & 3.17 & 2.14 & 18.25 &	8.20 &	14.65  \\
Sampled (15K) &  w/ \ourmethod{} (GPT)        & \textbf{11.06}$^\uparrow$ & \textbf{4.03}$^\uparrow$ & \textbf{8.20}$^\uparrow$ &	\textbf{8.34}$^\uparrow$ &	\textbf{3.51}$^\uparrow$ &	\textbf{6.54}$^\uparrow$ & 34.26$^\downarrow$ &	\textbf{8.80}$^\uparrow$ &	\textbf{21.70}$^\uparrow$  \\
Sampled (15K) &  w/ \ourmethod{} (Qwen)       & 10.51$^\uparrow$ &	3.11$^\uparrow$ &	7.12$^\uparrow$ &	7.60$^\uparrow$ & 2.55$^\uparrow$ &	5.44$^\uparrow$ & 32.66$^\downarrow$ &	7.70$^\uparrow$ &	19.81$^\uparrow$ \\
                                        \midrule 
 Sampled (15K)  & BART-Large          & 7.07 &	2.34 &	5.04 &	5.59 &	2.15 &	4.24  & 29.45 &	5.96 &	16.46      \\
 Full (156K) & BART-Large        & \textbf{11.08}	& 4.07 & \textbf{8.24} & \textbf{9.39} &	4.05 &	\textbf{7.43} & \textbf{39.34} &	9.01 &	\textbf{23.52}          \\ 
 Sampled (15K)  &  w/ \ourmethod{} (GPT)         & 10.11$^\uparrow$	& \textbf{4.48}$^\uparrow$ & 8.08$^\uparrow$ &	8.64$^\uparrow$ &	\textbf{4.49}$^\uparrow$ &	7.29$^\uparrow$ & 30.00$^\uparrow$ &	\textbf{9.50}$^\uparrow$ &	20.97$^\uparrow$          \\
 Sampled (15K)  &  w/ \ourmethod{} (Qwen)       & 8.46$^\uparrow$ & 3.52$^\uparrow$ & 6.60$^\uparrow$ & 7.06$^\uparrow$ &	3.41$^\uparrow$ & 5.81$^\uparrow$ & 31.22$^\uparrow$ & 5.99$^\uparrow$ &	16.94$^\uparrow$         \\
                                        
\midrule
Sampled (15K) & GECToR-Chinese              & 2.40 & 0.11 &	0.46 &	3.82 &	0.19 &	0.80 & 26.31 & 3.08 &	10.48         \\
Full (156K) & GECToR-Chinese          & 8.53 &	1.12 &	3.67 &	4.22 &	0.93 &	2.47 & 27.89 &	3.23 &	11.03  \\
Sampled (15K) &  w/ \ourmethod{} (GPT)         & \textbf{12.08}$^\uparrow$ & 2.19$^\uparrow$ &	6.35$^\uparrow$ &	\textbf{9.26}$^\uparrow$ &	1.87$^\uparrow$ &	5.17$^\uparrow$ & 30.55$^\uparrow$ & 4.74$^\uparrow$ & 14.62$^\uparrow$  \\
Sampled (15K) &  w/ \ourmethod{} (Qwen)         & 11.09$^\uparrow$ &	\textbf{2.63}$^\uparrow$ &	\textbf{6.74}$^\uparrow$ & 9.01$^\uparrow$ & \textbf{1.96}$^\uparrow$ & \textbf{5.24}$^\uparrow$ & \textbf{31.35}$^\uparrow$ &	\textbf{5.01}$^\uparrow$ & \textbf{15.28}$^\uparrow$ \\                         
\bottomrule
\end{tabular}
\caption{
Performance of various models on the NaCGEC benchmark. Note that 15K and 156K represent the amount of HSK data. $^\uparrow$ means that \ourmethod{} has improved performance compared to the baselines with the same training data.}
\label{Table: Main_Results_NaCGEC}
\end{table*}

\end{document}